%% file: main.tex
\documentclass[11pt,a4paper]{article}
\usepackage[hyperref]{naaclhlt2019}
\usepackage{times}
\usepackage{latexsym}
\usepackage{booktabs}
\usepackage{bm}
\usepackage{mathtools}
\usepackage{subcaption}
\usepackage{amsfonts}
\usepackage{subcaption}
\usepackage{pgfplots}
\usepackage{tikz}

\usepackage{multirow}

\usepackage{url}

\aclfinalcopy 
\def\aclpaperid{4022} 

\title{Learning When Not to Answer: \\ 
A Ternary Reward Structure for Reinforcement \\ 
Learning based Question Answering}

\author{
  Fr\'{e}deric Godin  \\
  Ghent University\\
  Ghent, Belgium \\
  \tt{ \small frederic.godin@ugent.be} \\
   \And
  Anjishnu Kumar \\
  Amazon Research \\
    Cambridge, United Kingdom\\
  \tt{\small anjikum@amazon.com} \\
   \And
  Arpit Mittal \\
  Amazon Research \\
  Cambridge, United Kingdom\\
  \tt{\small mitarpit@amazon.co.uk} \\ }

\date{}

\begin{document}
\maketitle
\begin{abstract}
In this paper, we investigate the challenges of using reinforcement learning agents for question-answering over knowledge graphs for real-world applications. We examine the performance metrics used by state-of-the-art systems and determine that they are inadequate for such settings. More specifically, they do not evaluate the systems correctly for situations when there is no answer available and thus agents optimized for these metrics are poor at modeling confidence. 
We introduce a simple new performance metric for evaluating question-answering agents that is more representative of practical usage conditions, and optimize for this metric by extending the binary reward structure used in prior work to a ternary reward structure which also rewards an agent for not answering a question rather than giving an incorrect answer. We show that this can drastically improve the precision of answered questions while only not answering a limited number of previously correctly answered questions.
Employing a supervised learning strategy using depth-first-search paths to bootstrap the reinforcement learning algorithm further improves  performance.
\end{abstract}

\section{Introduction}
\begin{figure}[t]
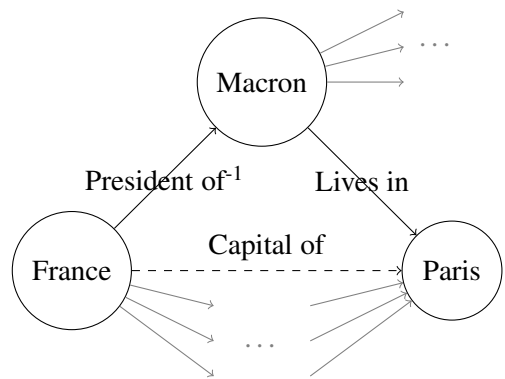

	\centering
    \include{ex_graph}
    \caption{Fictional graph for the the question ``What's the capital of France?". The relation $(Capital\ of)$ does not exist in the graph and thus an alternative path needs to be used that leads to the correct answer.}
    \label{fig:ex_graph}
\end{figure}

A number of approaches for question answering have been proposed recently that use reinforcement learning to reason over a knowledge graph \citep{minerva,LinRX2018:MultiHopKG,N18-1165,DBLP:conf/aaai/ZhangDKSS18}. In these methods the input question is first parsed into a constituent question entity and relation. The answer entity is then identified by sequentially taking a number of steps (or `hops') over the knowledge graph (KG) starting from the question entity.  
The agent receives a positive reward if it arrives at the correct answer entity and a negative reward for an incorrect answer entity. 
For example, for the question ``What is the capital of France?", the question entity is $(France)$ and the goal is to find a path in the KG which connects it to $(Paris)$. The relation between the answer entity and question entity in this example is $(Capital\ of)$ which is missing from the KG and has to be inferred via alternative paths. This is illustrated in Figure~\ref{fig:ex_graph}. A possible two-hop path to find the answer is to use the fact that $(Macron)$ is the president of $(France)$ and that he lives in $(Paris)$. However, there are many paths that lead to the entity $(Paris)$ but also to other entities which makes finding the correct answer a non-trivial task.

The standard evaluation metrics used for these systems are metrics developed for web search such as Mean Reciprocal Rank (MRR) and hits@k, where $k$ ranges from 1 to 20. We argue that this is not a correct evaluation mechanism for a practical question-answering system (such as Alexa, Cortana, Siri, etc.) where the goal is to return  a single  answer for each question. Moreover it is assumed that there is always an answer entity that could be reached from the question entity in limited number of steps. However this cannot be guaranteed in a large-scale commercial setting and for all KGs. For example, in our proprietary dataset used for the experimentation, for 15.60\% of questions the answer entity cannot be reached within the limit of number of steps used by the agent.
Hence, we propose a new evaluation criterion, 
allowing systems to return `no answer' as a response when no answer is available. 

We demonstrate that existing state-of-the-art methods are not suited for a practical question-answering setting and perform poorly in our evaluation setup. The root-cause of poor performance is the reward structure which does not provide any incentive to learn not to answer. The modified reward structure we present allows agents to learn not to answer in a principled way. Rather than having only two rewards, a positive and a negative reward, we introduce a ternary reward structure that also rewards agents for not answering a question.
A higher reward is given to the agent for correctly answering a question compared to not answering a question. 
In this setup the agent learns to make a trade-off between these three possibilities to obtain the highest total reward over all questions.

Additionally, because the search space of possible paths exponentially grows
with the number of hops, we also investigate using Depth-First-Search (DFS) algorithm to collect paths that lead to the correct answer. We use these paths as a supervised signal for training the neural network before the reinforcement learning algorithm is applied. We show that this improves overall performance.

\section{Related work}

The closest works to ours are the works by \citet{LinRX2018:MultiHopKG}, \citet{DBLP:conf/aaai/ZhangDKSS18} and \citet{minerva}, which consider the question answering task in a reinforcement learning setting in which the agent always chooses to answer.\footnote{An initial version of this paper has been presented at the Relational Representation Learning Workshop at NeurIPS 2018 as \citet{r2l_paper}.} Other approaches consider this as a link prediction problem in which multi-hop reasoning can be used to learn relational paths that link two entities. One line of work focuses on composing embeddings \cite{P15-1016,D15-1038,P16-1136} initially introduced for link prediction, 
e.g., TransE \cite{bordes2013translating}, ComplexE \cite{trouillon2016complex} or ConvE \cite{dettmers2018conve}. 
Another line of work focuses on logical rule learning such as neural logical programming \cite{DBLP:conf/nips/YangYC17} and neural theorem proving \cite{rocktaschel2017end}.
Here, we focus on question answering rather than link prediction or rule mining and use reinforcement learning to circumvent that we do not have ground truth paths leading to the answer entity.

Recently, popular textual QA datasets have been extended with not-answerable questions \citep{W17-2623,P18-2124}. Questions that cannot be answered are labeled with `no answer' option which allows for supervised training. This is different from our setup in which there are no ground truth `no answer' labels.

\begin{figure*}[t]
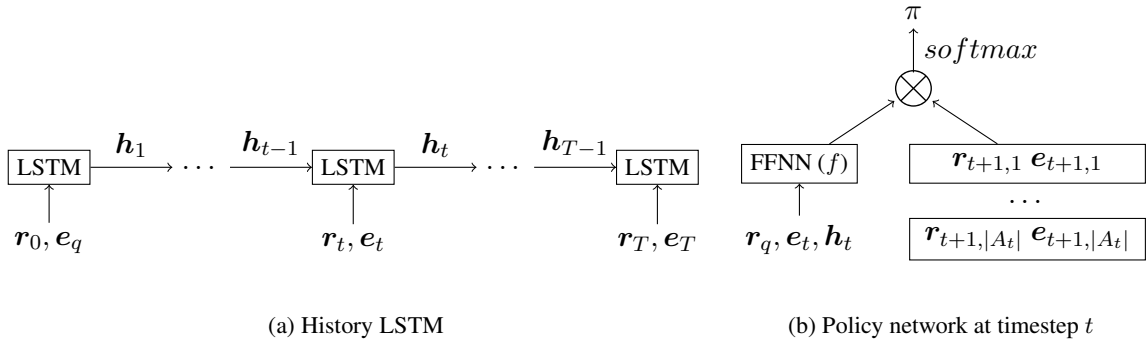

    \centering
    \begin{subfigure}[b]{0.6\textwidth}
        \centering
        \include{arch}
        \caption{History LSTM}
        \label{fig:arch_lstm}
    \end{subfigure}
	\centering
	    \begin{subfigure}[b]{0.35\textwidth}
        \centering
        \include{arch_policy}
        \caption{Policy network at timestep $t$}
        \label{fig:arch_policy}
    \end{subfigure}
	\centering
    
    \caption{Figure \ref{fig:arch_lstm} illustrates the LSTM which encodes  history of the path taken. The output at timestep $t$ is used as input to the policy network, illustrated in Figure \ref{fig:arch_policy}, to determine which action to take next.}
    \label{fig:arch}
\end{figure*}

\section{Background: Reinforcement learning}
We base our work on the recent reinforcement learning approaches introduced in \citet{minerva} and \citet{LinRX2018:MultiHopKG}. We denote the knowledge graph as $\mathcal{G}$, the set of entities as $\mathcal{E}$, the set of relations as $\mathcal{R}$ and the set of directed edges $\mathcal{L}$ between entities of the form $l = (e_1,r,e_2)$ with $e_1,e_2 \in \mathcal{E}$ and  $r \in \mathcal{R}$. 
The goal is to find an answer entity $e_a$ given a question entity $e_q$ and the question relation $r_q$, when $(e_q,r_q,e_a)$ is not part of graph $\mathcal{G}$. 

We formulate this problem as  a Markov Decision Problem (MDP) \citep{Sutton:1998:IRL:551283} with the following states, actions, transition function and rewards:  

\paragraph{States.} At every timestep $t$, the state $s_t$ is defined by the current entity $e_t$, the question entity $e_q$ and relation $r_q$, for which $e_t,e_q \in \mathcal{E}$ and $r_q \in \mathcal{R}$. More formally, $s_t = (e_t,e_q,r_q)$.

\paragraph{Actions.} For a given entity $e_t$, the set of possible actions is defined by the outgoing edges from $e_t$. Thus ${A}_t = \{(r',e')|(e_t,r',e') \in \mathcal{G}\}$.

\paragraph{Transition function.} The transition function $\delta$ maps $s_t$ to a new state $s_{t+1}$ based on the action taken by the agent. Consequently, $s_{t+1}=\delta(s_t,A_t)=\delta(e_t,e_q,r_q,A_t)$. 

\paragraph{Rewards.} The agent is rewarded based on the final state. For example, in \citet{minerva} and \citet{LinRX2018:MultiHopKG} the agent obtains a reward of 1 if the correct answer entity is reached as the final state and 0 otherwise (i.e., $R(s_T)=\mathbb{I}\{e_T=e_a\}$).

\subsection{Training}
We train a policy network $\pi$ using the REINFORCE algorithm of \citet{Williams:1992:SSG:139611.139614} which maximizes the expected reward:
\begin{equation}
    J(\bm{\theta}) = \mathbb{E}_{(e_q,r_q,e_a) \in \mathcal{G}} \mathbb{E}_{a_1,\dots,a_T \sim \pi} [R(s_T|e_q,r_q)]
\end{equation}
in which $a_t$ is the action selected at timestep $t$ following the policy $\pi$, and $\bm{\theta}$ are the parameters of the network.

The policy network consists of two parts: a Long Short-Term Memory (LSTM) network which encodes the history of the traversed path, and a feed-forward neural network to select an action ($a_t$) out of all possible actions. Each entity and relation have a corresponding vector $\bm{e}_t, \bm{r}_t \in \mathbb{R}^d$. The action $a_t \in A_t$ is represented by the vectors of the relation and entity as $\bm{a}_t = [\bm{r}_{t+1}; \bm{e}_{t+1}] \in \mathbb{R}^{2d}$. The LSTM encodes the history of the traversed path and updates its hidden state each timestep, based on the selected action:
\begin{equation}
    \bm{h}_t = LSTM(\bm{h}_{t-1},\bm{a}_{t-1})
\end{equation}
This is illustrated in Figure~\ref{fig:arch_lstm}.

Finally, the feed-forward neural network ($f$) combines the history $\bm{h}_t$, the current entity representation $\bm{e}_t$ and the query relation $\bm{r}_{q}$. Using softmax, we compute the probability for each action by calculating the dot product between the output of $f$ and each action vector $\bm{a}_t$:
\begin{equation}
    \pi(a_t|s_t) = softmax(\bm{A}_t  f(\bm{h}_t, \bm{e}_t, \bm{r}_{q}))
\end{equation}
in which $\bm{A}_t \in \mathbb{R}^{|A_t| \times 2d}$ is a matrix consisting of rows of action vectors $\bm{a}_t$. This is illustrated in Figure~\ref{fig:arch_policy}.
During training, we sample over this probability distribution to select the action $a_t$, whereas during inference, we use beam search to select the most probable path.

\section{Evaluation}
User-facing question answering systems inherently face a trade-off between presenting an answer to a user that could potentially be incorrect, and choosing not to answer. However, prior work in knowledge graph question-answering (QA) only considers cases in which the answering agent always produces an answer.
This setup originates from the link prediction and knowledge base completion tasks in which the evaluation criteria are hits@k and Mean Reciprocal Rank (MRR), where $k$ ranges from 1 to 20. However, these metrics are not an accurate representation of practical question-answering systems in which the goal is to return a single correct answer or not answer at all. Moreover, using these metrics result in the problem of the model learning  `spurious' paths since the metrics encourage the models 
to make wild guesses even if the path is unlikely to lead to the correct answer. 

We therefore propose to measure the fraction of questions the system answers (Answer Rate) and the number of correct answers out of all answers (Precision) 
to measure the system performance. We combine these two metrics by taking the harmonic mean and call this the QA Score. 
This can be viewed as a variant of the popular F-Score metric, with answer rate used as an analogue to recall in the original metric. 

\section{Proposed method} \label{sec:method}
In this section, we will first introduce the supervised learning technique we used to pretrain the neural network before applying the reinforcement learning algorithm. Next we will describe the ternary reward structure.

\subsection{Supervised learning}
Typically in reinforcement learning, the search space of possible actions and paths grows exponentially with the path length. Our problem is no exception to this. Hence an imitation learning approach could be beneficial here where we provide a number of expert paths to the learning algorithm to bootstrap the learning process. This idea has been explored previously in the context of link and fact prediction in knowledge graphs where \citet{D17-1060} proposed to use a Breadth-First-Search (BFS) between the entity pairs to select a set of plausible paths.
However BFS favours identification of shorter paths which could bias the learner. We therefore use Depth-First-Search (DFS) to identify paths between question and answer entities and sample up to 100 paths to be used for the supervised training. If no path can be found between the entity pair we return a `no answer' label. 
Following this, we train the network using reinforcement learning algorithm which refines it further. 
Note that it is not guaranteed that the set of paths found using DFS are all most efficient. However as we show in our experiments, bootstrapping with these paths provide good initialization for the reinforcement learning algorithm.

\subsection{Ternary reward structure}\label{sec:ternary}

\begin{figure}[t]
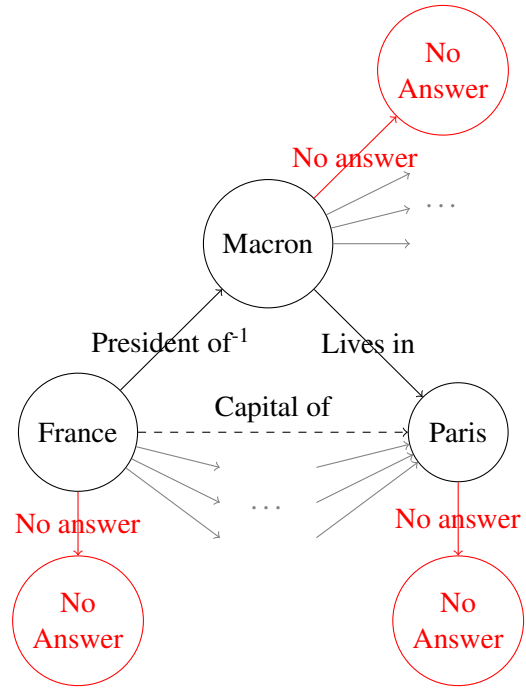

	\centering
    \include{ex_graph_noanswer}
    \caption{Fictional graph for the the question ``What's the capital of France?". The relation $(Capital\ of)$ does not exist in the graph and thus an alternative path needs to be used that leads to the correct answer. To avoid that the agent returns an incorrect answer when not finding the correct answer, a `no answer' relation is added between every entity node and a special `no answer' node, to be able to return `no answer'. }
    \label{fig:ex_graph_noanswer}
\end{figure}

As mentioned previously, we encounter situations when the answer entity cannot be reached in the limited number of steps taken by an agent. In such cases, the system should return a special answer `no answer' as the response. We can achieve this by adding a synthetic `no answer' action that leads to a special entity $e_{NO ANSWER}$. This is illustrated in Figure~\ref{fig:ex_graph_noanswer}.
In the framework of \citet{minerva} a binary reward is used which rewards the learner for the answer being wrong or correct. Following a similar protocol, we could award a score of 1 to return `no answer' when there is no answer available in the KG. However, we cannot achieve reasonable training with such reward structure. This is because there is no specific pattern for `no answer' that could be directly learned. Hence, if we reward a system equally for correct or no answer, it learns to always predict `no answer'. 
We therefore propose a ternary reward structure in which a positive reward is given to a correct answer, a neutral reward when $e_{NO ANSWER}$ is selected as an answer, and a negative reward for an incorrect answer. More formally:
\begin{equation}
    R(s_T) =   
    \begin{cases} 
   r_{pos} & \text{if } e_T=e_a, \\
   0       & \text{if } e_T=e_{NO ANSWER}, \\
   r_{neg} & \text{if } e_T \not\in \{e_a,e_{NO ANSWER}\} \\
  \end{cases}
\end{equation}
with $r_{pos} \textgreater 0$ and $r_{neg} \textless 0$. The idea is that the agent receives a larger reward for a correct answer compared to not answering the question, and a negative reward for incorrectly answering a question compared to not answering the question. In the experimental section, we show that this mechanism provides better performance.

\begin{table*}[t]
\caption{Results on FB15k-237 dataset.}
\centering
\input{results_table.tex}
\label{tab:results_fb}

\end{table*}

\begin{table*}[t]
\caption{Results on Alexa69k-378 dataset.}
\centering
\input{results_table_evi.tex}
\label{tab:results_evi}

\end{table*}

\begin{table}[t]
\caption{Statistics of various datasets.}
\centering
\input{datasets.tex}
\label{tab:dataset}

\end{table}

\section{Experimental setup}
We evaluate our proposed approach on a publicly available dataset, FB15k-237 \citep{fb15k237} which is based on the Freebase knowledge graph and a proprietary dataset Alexa69k-378 which is a sample of Alexa's proprietary knowledge graph. Both the public dataset and the proprietary dataset are good examples of real-world general-purpose knowledge graphs that can be used for question answering. FB15k-237 contains 14,505 different entities and 237 different relations resulting in 272,115 facts. Alexa69k-378 contains 69,098 different entities and 378 different relations resulting in 442,591 facts. We follow the setup of \citet{minerva}, using the same train/val/test splits for FB15k-237. For Alexa69k-378 we use 10\% of the full dataset for validation and test. For both datasets, we add the reverse relations of all relations in the training set in order to facilitate backward navigation following the approach of previous work. Similarly, a `no op' relation is added for each entity between the entity and itself, which allows the agent to loop/reason multiple consecutive steps over the same entity.  An overview of both datasets can be found in Table~\ref{tab:dataset}.

We extend the publicly available implementation of \citet{minerva} for our experimentation. 
We set the size of the entity and relation representations $d$ at 100 and the hidden state at 200. We use a single layer LSTM and train models with path length 3 (tuned using hyper-parameter search). We optimize the neural network using Adam \citep{DBLP:journals/corr/KingmaB14} with learning rate 0.001, mini-batches of size 256 with 20 rollouts per example. During the test time, we use beam search with the beam size of 100. Unlike \citet{minerva}, we also train entity embeddings after initializing them with random values. 
Reward values are set as $r_{pos}=10$ and $r_{neg}=-0.1$ after performing a coarse grid search for various reward values on the validation set.  For all experiments, we selected the best model with the highest QA Score on the corresponding validation set.

\section{Results} \label{sec:results}

The results of our experiments for FB15k-237 and Alexa69k-378 are given in Table~\ref{tab:results_fb} and Table~\ref{tab:results_evi} respectively. 
\paragraph{Supervised learning}
For FB15k-237, we see that the model trained using reinforcement learning (RL) scores as well as the model trained using supervised learning. This makes supervised learning using DFS a strong baseline system for question answering over knowledge graphs, and for FB15k-237 in particular. On Alexa69k-378, models trained using supervised learning score lower on all metrics compared to RL. When combining supervised learning with RL overall performance increases.

\paragraph{No answer}
When we train RL system with our ternary reward structure (No Answer RL), the precision and QA score increase significantly on both datasets. For FB15k-237, our No Answer RL model decided not to answer over 40\% of the questions, with an absolute hits@1 reduction of only 1.3\% over standard RL. Moreover, of all the answered questions, 40.11\% were answered correctly compared to 24.75\% of the original question-answering system: an absolute improvement of over 15\%. This resulted in the final QA Score of 47.58\%, around 8\% higher than standard RL and 12\% higher than \citet{minerva}.

Similarly, 60\% of the questions did not get answered on Alexa69k-378. This resulted in hits@1 decrease of roughly 1\% but compared to standard RL, the precision increased from 16.77\% to 38.92\%: an absolute increase of more than 20\%. The final QA Score also increased from 28.72\% to 39.55\%, and also significantly improved over \citet{minerva} and \citet{LinRX2018:MultiHopKG}. 
The results indicate that using our method allows us to improve the precision of the question-answering system by choosing the right questions to be answered by not answering many questions that were previously answered incorrectly. This comes at the expense of not answering some questions that previously could be answered correctly.

\paragraph{All}
Finally, all methods were combined in a single method. First the model was pretrained in a supervised way. 
Then the model was retrained using RL algorithm with ternary reward structure. This jointly trained model obtained better QA scores than any individually trained model. On FB15k-237, a QA score of 52.16\% is obtained which is an absolute improvement of 4.58\% over the best individual model and 2.66\% over \citet{LinRX2018:MultiHopKG}. Similarly, on Alexa69k-378, an absolute improvement of 2.57\% over the best individual result is obtained, almost 10\% absolute improvement over \citet{LinRX2018:MultiHopKG}.
Sample results from our method are given in Table~\ref{tab:ex_correct} and Table~\ref{tab:ex_incorrect}. 

\begin{table*}[t]
\caption{Example paths of correctly answered questions on FB15k-237. Note that the fact $(e_q,r_q,e_a)$ is not part of the KG.}
\centering
\input{examples.tex}
\label{tab:ex_correct}
\end{table*}

\begin{table*}[t]
\caption{Example question from FB15k-237, incorrectly answered by \cite{minerva} and not answered by our system. Note that the fact $(e_q,r_q,e_a)$ is not part of the KG. }
\centering
\input{examples_incorrect.tex}
\label{tab:ex_incorrect}
\end{table*}

\begin{figure}[t]
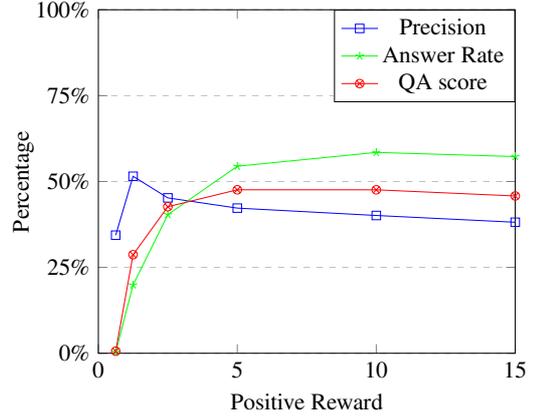

	\centering
    \include{positive_reward}
    \caption{Influence of changing the positive reward for FB15k-237. The negative reward is fixed at $r_{neg}=-0.1$ and the neutral reward is fixed at $r_{neutral}=0$.}
    \label{fig:pos_reward}
\end{figure}
\begin{figure}[t]
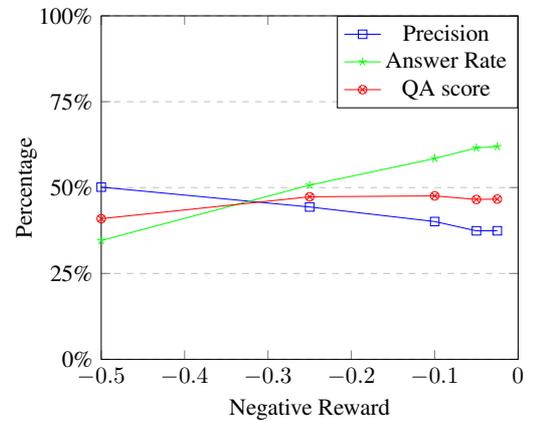

	\centering
    \include{negative_reward}
    \caption{Influence of changing the negative reward for FB15k-237. The positive reward is fixed at $r_{pos}=10$ and the neutral reward is fixed at $r_{neutral}=0$.}
    \label{fig:neg_reward}
\end{figure}

\paragraph{Reward tuning}
An important part of increasing the QA score is to select the right combination of rewards. Therefore, we ran additional experiments where we varied either the positive or negative reward, keeping the other rewards fixed. 
In Figure~\ref{fig:pos_reward}, the precision, answer rate and QA score are shown when varying the positive reward and keeping the neutral and negative rewards fixed. When, the positive reward is very small ($r_{pos}=0.625$), almost no question is answered. When the positive reward $r_{pos}$ is $1.25$, roughly 20\% of the questions are answered with a 50\% precision. After that, the precision starts declining and the answer rate starts increasing, resulting in an overall increase in QA score. The QA score plateaus between 5 and 10 and then starts decreasing slowly. In Figure~\ref{fig:neg_reward}, the precision, answer rate and QA score are shown when varying the negative reward and keeping the neutral and positive rewards fixed. In this case, the highest QA score is achieved when the negative reward is between -0.25 and -0.1. 
As long as the negative reward is lower than zero, a wrong answer gets penalized and the QA score stays high.

\section{Conclusions}
In this paper, we addressed the limitations of current approaches for question answering over a knowledge graph that use reinforcement learning. 
Rather than only returning a correct or incorrect answer, we allowed the model to not answer a question when it is not sure about it. Our ternary reward structure gives different rewards for correctly answered, incorrectly answered and not answered questions. We also introduced a new evaluation metric which takes these three options into account. We showed that we can significantly improve the precision of answered questions compared to previous approaches, making this a promising direction for the practical usage in knowledge graph-based QA systems.

\bibliographystyle{acl_natbib}

\end{document}

%% file: ex_graph.tex
\begin{tikzpicture}

\node at (0,0) [circle,draw,inner sep=+0.5em] (h1) {France};
\node at (2,-0.5) [] (h7) {};
\node at (2,-1.) [] (h8) {};
\node at (2,-1.5) [] (h9) {};
\node at (2.5,-1) [gray] () {\dots};
\node at (5,0) [circle,draw,inner sep=+0.5em] (h2) {Paris};
\node at (3,-0.5) [] (h13) {};
\node at (3,-1.) [] (h14) {};
\node at (3,-1.5) [] (h15) {};

\node at (2.5,2.5) [circle,draw,inner sep=+0.5em] (h3) {Macron};
\node at (4.5,2.5) [] (h4) {};
\node at (4.5,3) [] (h5) {};
\node at (4.5,3.5) [] (h6) {};
\node at (4.8,3) [gray] () {\dots};

\draw[->,dashed] (h1) -- node[above]  {Capital of}  (h2) ;
\draw[->,gray] (h1) -- (h7) ;
\draw[->,gray] (h1) -- (h8) ;
\draw[->,gray] (h1) -- (h9) ;
\draw[->] (h1) -- node {President of\textsuperscript{-1}}  (h3) ;
\draw[->,gray] (h13) -- (h2) ;
\draw[->,gray] (h14) -- (h2) ;
\draw[->,gray] (h15) -- (h2) ;
\draw[->] (h3) -- node  {Lives in}  (h2) ;
\draw[->,gray] (h3) -- (h4) ;
\draw[->,gray] (h3) -- (h5) ;
\draw[->,gray] (h3) -- (h6) ;

\end{tikzpicture}

%% file: arch.tex
\begin{tikzpicture}
\node at (0,0) [rectangle,draw,inner sep=+0.3em] (h1) {\small LSTM};
\node at (2,0) [inner sep=+0.3em] (h3) {$\dots$};
\node at (4,0) [rectangle,draw,inner sep=+0.3em] (h4) {\small LSTM};
\node at (6,0) [inner sep=+0.3em] (h5) {$\dots$};
\node at (8,0) [rectangle,draw,inner sep=+0.3em] (h6) {\small LSTM};
\draw[->] (h1) -- node[above]  {$\bm{h}_1$}  (h3) ;
\draw[->] (h3) -- node[above]  {$\bm{h}_{t-1}$} (h4) ;
\draw[->] (h4) -- node[above]  {$\bm{h}_t$} (h5) ;
\draw[->] (h5) -- node[above]  {$\bm{h}_{T-1}$} (h6) ;

\node at (0,-1) [inner sep=+0.3em] (x1) {$\bm{r}_0, \bm{e}_q$};

\node at (4,-1) [inner sep=+0.3em] (x4) {$\bm{r}_t, \bm{e}_t$};

\node at (8,-1) [inner sep=+0.3em] (x6) {$\bm{r}_T, \bm{e}_T$};
\draw[->] (x1) -- (h1) ;

\draw[->] (x4) -- (h4) ;
\draw[->] (x6) -- (h6) ;

\end{tikzpicture}

%% file: arch_policy.tex
\begin{tikzpicture}
\node at (0,0) [rectangle,draw,inner sep=+0.3em] (h1) {\small FFNN ($f$)};
\node at (0,-1) [inner sep=+0.3em] (x1) {$\bm{r}_q, \bm{e}_t, \bm{h}_t$};
\draw[->] (x1) -- (h1) ;

\node at (3,0) [rectangle,draw,inner sep=+0.3em,minimum width=3.1cm] (a1) {$\bm{r}_{t+1,1}\  \bm{e}_{t+1,1}$};
\node at (3,-0.5) [inner sep=+0.3em] (a2) {$\dots$};
\node at (3,-1) [rectangle,draw,inner sep=+0.3em,minimum width=3.1cm] (a3) {$\bm{r}_{t+1,|A_t|}\ \bm{e}_{t+1,|A_t|}$};

\node at (1.5,1) [inner sep=-0.1em] (d) {\huge $\otimes$};

\draw[->] (h1) -- (d) ;
\draw[->] (a1) -- (d) ;

\node at (1.5,2) [inner sep=+0.3em] (out) {$\pi$};
\draw[->] (d) -- node[right]  {$softmax$} (out) ;

\end{tikzpicture}

%% file: ex_graph_noanswer.tex
\begin{tikzpicture}

\node at (0,0) [circle,draw,inner sep=+0.5em] (h1) {France};
\node at (2,-0.5) [] (h7) {};
\node at (2,-1.) [] (h8) {};
\node at (2,-1.5) [] (h9) {};
\node at (2.5,-1) [gray] () {\dots};
\node at (0,-2.5) [circle,draw,inner sep=+0.3em,red,text width=1.2cm,align=center] (h11) {No Answer};
\node at (5,0) [circle,draw,inner sep=+0.5em] (h2) {Paris};
\node at (3,-0.5) [] (h13) {};
\node at (3,-1.) [] (h14) {};
\node at (3,-1.5) [] (h15) {};
\node at (5,-2.5) [circle,draw,inner sep=+0.3em,red,text width=1.2cm,align=center] (h12) {No Answer};

\node at (2.5,2.5) [circle,draw,inner sep=+0.5em] (h3) {Macron};
\node at (4.5,2.5) [] (h4) {};
\node at (4.5,3) [] (h5) {};
\node at (4.5,3.5) [] (h6) {};
\node at (4.8,3) [gray] () {\dots};
\node at (4.8,4.8) [circle,draw,inner sep=+0.3em,red,text width=1.2cm,align=center] (h10) {No Answer};

\draw[->,dashed] (h1) -- node[above]  {Capital of}  (h2) ;
\draw[->,gray] (h1) -- (h7) ;
\draw[->,gray] (h1) -- (h8) ;
\draw[->,gray] (h1) -- (h9) ;
\draw[->,red] (h1) -- node  {No answer}  (h11) ;
\draw[->,red] (h2) -- node  {No answer}  (h12) ;
\draw[->] (h1) -- node {President of\textsuperscript{-1}}  (h3) ;
\draw[->,gray] (h13) -- (h2) ;
\draw[->,gray] (h14) -- (h2) ;
\draw[->,gray] (h15) -- (h2) ;
\draw[->] (h3) -- node  {Lives in}  (h2) ;
\draw[->,gray] (h3) -- (h4) ;
\draw[->,gray] (h3) -- (h5) ;
\draw[->,gray] (h3) -- (h6) ;
\draw[->,red] (h3) -- node  {No answer} (h10) ;

\end{tikzpicture}

%% file: results_table.tex
 \begin{tabular}{c c c c | c c c} 
\toprule
Model & Hits@1 &  Hits@10 & MRR & Precision & Answer Rate & QA Score \\ 
\midrule

\cite{minerva} & 0.217 & 0.456 & 0.293 & 0.217 & \textbf{1} & 0.357\\
\cite{LinRX2018:MultiHopKG} & 0.329 & 0.544 & 0.393 & 0.329 & \textbf{1} & 0.495\\
\midrule
RL & 0.2475 & 0.4032 & 0.2983 & 0.2475 & \textbf{1} &	0.3968 \\ 
Supervised & 0.2474 & 0.4929 & 0.3276 & 0.2474 & \textbf{1} &	0.3967 \\ 
Supervised + RL & 0.2736 & 0.5015 & 0.3469 & 0.2736 & \textbf{1} & 0.4296 \\
No Answer RL & 0.2345 & 0.3845 & 0.2831 & 0.4011 & 0.5847 & 0.4758 \\
\midrule
All & 0.2738 & 0.4412 & 0.3286 & \textbf{0.4835} & 0.5663 & \textbf{0.5216} \\

 \bottomrule
 \end{tabular}

%% file: results_table_evi.tex
 \begin{tabular}{c c c c | c c c} 
\toprule
Model & Hits@1 &  Hits@10 & MRR & Precision & Answer Rate & QA Score \\ 
\midrule
\cite{minerva} & 0.1790 & 0.2772 & 0.2123 &  0.1790 & \textbf{1} & 0.3036\\
\cite{LinRX2018:MultiHopKG} & 0.1915 & 0.3184 & 0.2358 & 0.1915 & \textbf{1} & 0.3214\\
\midrule
RL &  0.1677 & 0.2716 & 0.2031 & 0.1677 & \textbf{1} & 0.2872\\ 
Supervised &  0.1471 & 0.3142 & 0.203 & 0.1471 & \textbf{1} & 0.2565 \\ 
Supervised + RL &  0.1937 & 0.3045 & 0.2312 & 0.1937 & \textbf{1} & 0.3245\\
No Answer RL & 0.1564 & 0.2442 & 0.1858 & \textbf{0.3892} & 0.4019 & 0.3955\\
\midrule
All & 0.1865 & 0.294 & 0.2229 & 0.3454 & 0.5401 & \textbf{0.4213}\\

 \bottomrule
 \end{tabular}

%% file: datasets.tex
 \begin{tabular}{c c c c c} 
\toprule
&&&\multicolumn{2}{c}{\#queries}\\
\cline{4-5}
\#ent & \#rel & \#facts & valid & test \\
\midrule
\multicolumn{5}{c}{FB15k-237}\\
\midrule
14,505 & 237 & 272,115 & 17,535 & 20,466\\
\midrule
\multicolumn{5}{c}{Alexa69k-378}\\
\midrule
69,098 & 378 & 442,591 & 55,186 & 55,474 \\
 \bottomrule
 \end{tabular}

%% file: examples.tex
\begin{tabular}{l}
\toprule
\begin{tikzpicture}
   \node[style={font=\small}] at (0, 1) (z) {Question: $e_q$ = Bruce Broughton, $r_q$ = Profession. Answer: $e_a$ = Music Composer};
\end{tikzpicture}\\
\begin{tikzpicture}
  \node[style={font=\small}] at (0, 0)   (a) {Bruce Broughton};
  \node[style={font=\small}] at (4, 0)   (b) {Oscar Best Music};
  \node[style={font=\small}] at (8, 0)   (c) {Nino Rota};
  \node[style={font=\small}] at (11, 0)   (d) {Music Composer};
  \draw[->,thick](a) -- node[above,style={font=\tiny}] {Award Nominee}   (b);
  \draw[->,thick](b) -- node[above,style={font=\tiny}] {Award Winner}   (c);
  \draw[->,thick](c) -- node[above,style={font=\tiny}] {Profession}   (d);
  
\end{tikzpicture}\\
\midrule
\begin{tikzpicture}
   \node[style={font=\small}] at (0, 1) (z) {Question: $e_q$ = Washington nationals, $r_q$ = Sports Team Sport. Answer: $e_a$ = Baseball};
\end{tikzpicture}\\
\begin{tikzpicture}
  \node[style={font=\small}] at (0, 0)   (a) {Washington Nationals};
  \node[style={font=\small}] at (4, 0)   (b) {National League};
  \node[style={font=\small}] at (8, 0)   (c) {Milwaukee Braves};
  \node[style={font=\small}] at (11, 0)   (d) {Baseball};
  \draw[->,thick](a) -- node[above,style={font=\tiny}] {Sports League\textsuperscript{-1}}   (b);
  \draw[->,thick](b) -- node[above,style={font=\tiny}] {Sports League}   (c);
  \draw[->,thick](c) -- node[above,style={font=\tiny}] {Sports Team Sport}   (d);
  
\end{tikzpicture}\\
\bottomrule
\end{tabular}

%% file: examples_incorrect.tex
\begin{tabular}{l}
\toprule
\begin{tikzpicture}
   \node[style={font=\small}] at (0, 1) (z) {Question: $e_q$ = Sherlock holmes (movie), $r_q$ = Story By. Answer: $e_a$ = Conan Doyle};
\end{tikzpicture}\\
\begin{tikzpicture}
  \node[style={font=\small}] at (0, 0)   (a) {Sherlock holmes (movie)};
  \node[style={font=\small}] at (5, 0)   (b) {Wardrobe Supervisor};
  \node[style={font=\small}] at (8, 0)   (c) {Wardrobe Sup.};
  \node[style={font=\small}] at (11, 0)   (d) {Wardrobe Sup.};
  \draw[->,thick](a) -- node[above,style={font=\tiny}] {Film Crew Role}   (b);
  \draw[->,thick](b) -- node[above,style={font=\tiny}] {No op}   (c);
  \draw[->,thick](c) -- node[above,style={font=\tiny}] {No op}   (d);
  
\end{tikzpicture}\\
\begin{tikzpicture}
  \node[style={font=\small}] at (0, 0)   (a) {Sherlock holmes (movie)};
  \node[style={font=\small}] at (5, 0)   (b) {Wardrobe Supervisor};
  \node[style={font=\small}] at (8, 0)   (c) {No answer};
  \node[style={font=\small}] at (11, 0)   (d) {No answer};
  \draw[->,thick](a) -- node[above,style={font=\tiny}] {Film Crew Role}   (b);
  \draw[->,thick](b) -- node[above,style={font=\tiny}] {No answer}   (c);
  \draw[->,thick](c) -- node[above,style={font=\tiny}] {No answer}   (d);
  
\end{tikzpicture}\\
\bottomrule
\end{tabular}

%% file: positive_reward.tex
\begin{tikzpicture}[scale=0.8]
\begin{axis}[
    xlabel={Positive Reward},
    ylabel={Percentage},
    xmin=0, xmax=15,
    ymin=0, ymax=1,
    xtick={0,5,10,15},
    xticklabels={0,5,10,15},
    ytick={0,0.25,0.5,0.75,1},
    yticklabels={0\%,25\%,50\%,75\%,100\%},
	legend style={at={(1,1)},anchor=north east},
    ymajorgrids=true,
    grid style=dashed
] 
\addplot[
    color=blue,
    mark=square,
    ]
    coordinates {
    (0.625,0.3438)(1.25,0.5152)(2.5,0.4523)(5,0.4226)(10,0.4011)(15,0.3814)
    };
    \addlegendentry{Precision}
\addplot[
    color=green,
    mark=star,
    ]
    coordinates {
    (0.625,0.0031)(1.25,0.1988)(2.5,0.4036)(5,0.5444)(10,0.5847)(15,0.5726)
    };
    \addlegendentry{Answer Rate}
\addplot[
    color=red,
    mark=otimes,
    ]
    coordinates {
    (0.625,0.0062)(1.25,0.2869)(2.5,0.4266)(5,0.4759)(10,0.4758)(15,0.4579)
    };
    \addlegendentry{QA score}
\end{axis}
\end{tikzpicture}

%% file: negative_reward.tex
\begin{tikzpicture}[scale=0.8]
\begin{axis}[
    xlabel={Negative Reward},
    ylabel={Percentage},
    xmin=-0.5, xmax=0,
    ymin=0, ymax=1,
    ytick={0,0.25,0.5,0.75,1},
    yticklabels={0\%,25\%,50\%,75\%,100\%},
	legend style={at={(1,1)},anchor=north east},
    ymajorgrids=true,
    grid style=dashed
] 
\addplot[
    color=blue,
    mark=square,
    ]
    coordinates {
    (-0.5,0.5013)(-0.25,0.4435)(-0.1,0.4011)(-0.05,0.3741)(-0.025,0.3742)
    };
    \addlegendentry{Precision}
\addplot[
    color=green,
    mark=star,
    ]
    coordinates {
    (-0.5,0.3459)(-0.25,0.5071)(-0.1,0.5847)(-0.05,0.6156)(-0.025,0.6194)
    };
    \addlegendentry{Answer Rate}
\addplot[
    color=red,
    mark=otimes,
    ]
    coordinates {
    (-0.5,0.4094)(-0.25,0.4732)(-0.1,0.4758)(-0.05,0.4654)(-0.025,0.4665)
    };
    \addlegendentry{QA score}
\end{axis}
\end{tikzpicture}